\pdfoutput=1
\documentclass{article}

\usepackage[preprint]{neurips_2023}

\usepackage[utf8]{inputenc} 
\usepackage[T1]{fontenc}    
\usepackage{hyperref}       
\usepackage{url}            
\usepackage{booktabs}       
\usepackage{amsfonts}       
\usepackage{nicefrac}       
\usepackage{microtype}      
\usepackage[table,xcdraw]{xcolor}         

\usepackage{subcaption}
\usepackage{wrapfig}
\usepackage{multirow}
\usepackage{makecell}
\usepackage{xspace}
\usepackage{enumitem}
\usepackage{amsmath}
\usepackage{listings}
\usepackage{algorithm}
\usepackage{algorithmic}
\usepackage{amsfonts,amssymb}
\usepackage{mathrsfs}
\usepackage{subcaption}
\usepackage{natbib}
\setcitestyle{numbers,square}
\usepackage{graphicx}       
\usepackage{cleveref}
 
\usepackage{inconsolata}
\usepackage{amsmath}
\usepackage{amssymb}
\usepackage{pgfplots}
\usepackage{multirow}
\usepackage{booktabs}
\usepackage{float}
\usepackage{xspace}
\usepackage{epigraph}

\usepackage{booktabs}
\usepackage{multirow}
\usepackage{enumitem}
\usepackage[table,xcdraw]{xcolor}
\usepackage[normalem]{ulem}
\useunder{\uline}{\ul}{}

\usepackage[framemethod=TikZ]{mdframed}

\newcommand{\hide}[1]{}
\newcommand{\model}[0]{\textsc{Self-Contrast}\xspace}
\newcommand{\vpara}[1]{\noindent\textbf{#1}\xspace} %

\newtheorem{definition}{Definition}
\newtheorem{remark}{Remark}
\newenvironment{pf}{{ \it{Proof.}}\quad}{}

\newmdtheoremenv[%
backgroundcolor=yellow!10,%
outerlinecolor=black,%
innertopmargin = \topskip,%
splittopskip = \topskip,%
ntheorem = false,%
roundcorner=4pt]
{framedtheorem}{Theorem}

\newmdtheoremenv[%
backgroundcolor=gray!10,%
outerlinecolor=black,%
innertopmargin = \topskip,%
splittopskip = \topskip,%
ntheorem = false,%
roundcorner=4pt]
{assumption}{Assumption}

\definecolor{lightlightgray}{HTML}{EFEFEF}

\graphicspath{{figures/}}

\title{Extensive Self-Contrast Enables Feedback-Free\\Language Model Alignment}

\author{Xiao Liu$^{12*}$, Xixuan Song$^{12\dagger*}$, Yuxiao Dong$^{2}$, Jie Tang$^{2}$ \\
$^1$Zhipu AI \quad\quad $^2$Tsinghua University
\\ \\
{\includegraphics[height=3.5ex]{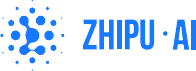}}
}

\begin{document}

\maketitle
\renewcommand{\thefootnote}{\fnsymbol{footnote}}
    \footnotetext[1]{XL and XS contributed equally. Emails: \texttt{shawliu9@gmail.com},\texttt{ songxx21@mails.tsinghua.edu.cn}}
    \footnotetext[2]{Work done while XS interned at Zhipu AI.}
\renewcommand{\thefootnote}{\arabic{footnote}}

\vspace{-8mm}
\begin{abstract}
Reinforcement learning from human feedback (RLHF) has been a central technique for recent large language model (LLM) alignment.
However, its heavy dependence on costly human or LLM-as-Judge preference feedback could stymie its wider applications.
In this work, we introduce \model, a feedback-free large language model alignment method via exploiting extensive self-generated negatives.
With only supervised fine-tuning (SFT) targets, \model leverages the LLM itself to generate massive diverse candidates, and harnesses a pre-trained embedding model to filter multiple negatives according to text similarity.
Theoretically, we illustrate that in this setting, merely scaling negative responses can still effectively approximate situations with more balanced positive and negative preference annotations.
Our experiments with direct preference optimization (DPO) on three datasets show that, \model could consistently outperform SFT and standard DPO training by large margins.
And as the number of self-generated negatives increases, the performance of \model continues to grow.
Code and data are available at \url{https://github.com/THUDM/Self-Contrast}. 
\end{abstract}

\section{Introduction}
\begin{figure*}[htb]
\centering
\hspace{-2mm}
\begin{minipage}[t]{.52\textwidth}
\centering

    \includegraphics[width=\textwidth]{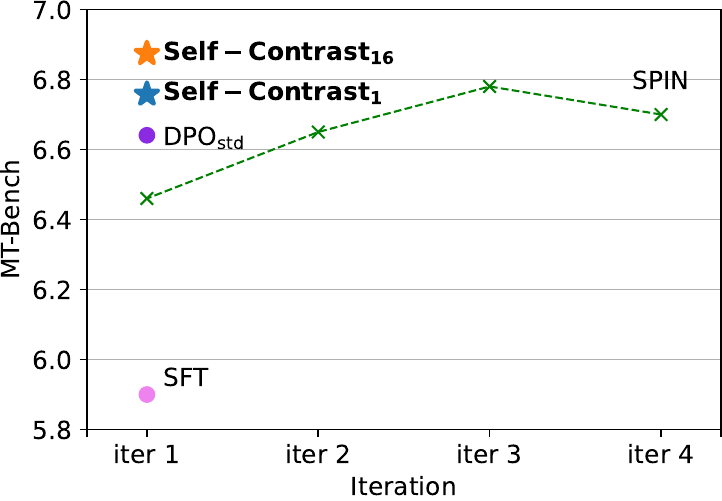}
    \caption{Feedback-free \model achieves higher MT-Bench scores with only SFT prompts, targets, and self-generated negative samples without iterative training compared to previous approaches.}
    \label{fig:Compare_with_SPIN}

\end{minipage}%
\hspace{1.5mm}
\begin{minipage}[t]{.46\textwidth}
\centering

    \includegraphics[width=\textwidth]{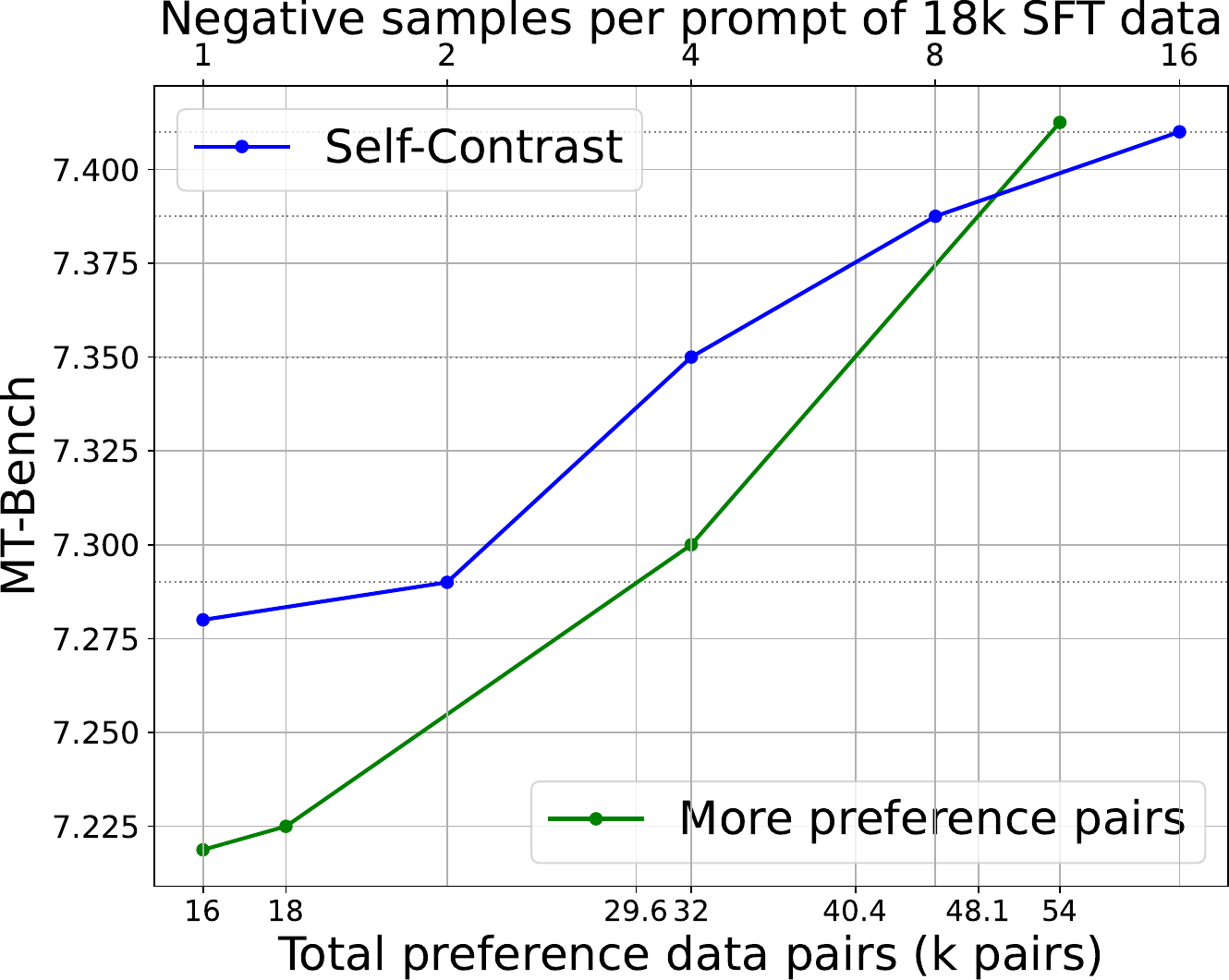}
    \caption{Scaling \model and standard preference pairs comparison. Feedback-free \model is competitive to training with $\times 3$ more preference feedback labels.}
    \label{fig:Compare_with_More}

\end{minipage}%
\end{figure*}

Large Language Models (LLMs) such as GPT-3~\cite{GPT3}, PALM~\cite{chowdhery2022palm}, OPT~\cite{zhang2022opt}, GLM~\cite{zeng2022glm,du2022glm}, and LLaMA~\cite{touvron2023llama} have made significant strides in producing outputs that are not only accurate but also meaningful and useful in human contexts. 
A critical aspect of directing these pre-trained LLMs towards understanding human intentions is the concept of alignment~\cite{ouyang2022training,bai2022constitutional}, primarily achieved through Supervised Fine-Tuning (SFT) and Reinforcement Learning from ``X'' Feedback (RLxF) stages. 
Specifically, RLxF techniques~\cite{schulman2017proximal,rafailov2023direct,zhao2023slichf,wang2023openchat}, which utilize human preferences to provide feedback on LLM responses, are seen as essential for enhancing LLM alignment.

However, a significant challenge in scaling RLxF lies in collecting preference feedback, which is often costly, whether obtained from humans~\cite{ouyang2022training} or advanced AI systems like GPT-4~\cite{openai2023gpt4}. 
Consequently, there has been a growing interest in feedback-minimal approaches to LLM alignment. 
While SFT has seen advancements in reducing human supervision through works like~\cite{sun2023principledriven,zhou2023lima}, less attention has been paid to feedback-free alignment during the RLHF phase, which could significantly contribute to the performance improvement of LLM alignment training. 
Recent efforts in feedback-free alignment, such as those by ~\cite{yuan2024selfrewarding,chen2024selfplay}, are noteworthy, yet their reliance on multiple iterations may limit efficiency in practice.

Addressing this challenge, we introduce a novel feedback-free LLM alignment method, \model, designed to bypass the need for labor-intensive preference comparisons. 
Our approach leverages the abundance of self-generated negatives, which we theorize can significantly contribute to the efficiency of RLHF training, especially under the assumption that negative responses are more varied than positive ones. 
We theoretically demonstrate that with a sufficient number of negative responses, albeit with fewer positives, we can effectively approximate the optimization effect achieved with balanced comparison pairs under certain conditions (Cf. Theorem~\ref{the:multi-negatives}).

Empirically, we implement \model across three preference datasets~\cite{starling2023,tunstall2023zephyr,bai2022constitutional} and employ Direct Preference Optimization (DPO,~\cite{rafailov2023direct}), utilizing SFT targets without the need for comparison feedback or iterative training. 
Our extensive experiments reveal that \model not only surpasses other feedback-free methods~\cite{chen2024selfplay} but also outperforms its DPO counterparts trained with original feedback dataset. 
Notably, our findings indicate that increasing the volume of self-generated negatives from 1 to 16 continues to enhance performance, particularly on the Nectar~\cite{starling2023} and HH-RLHF$_{test}$~\cite{bai2022constitutional} datasets, underscoring the untapped potential of \model in elevating LLM alignment efficacy. 
Further, our quantitative analysis corroborates the theoretical advantages of employing self-generated negatives over traditional balanced comparison pairs.

In summary, our contributions are as follows:
\begin{itemize}[leftmargin=1.5em,itemsep=0pt,parsep=0.2em,topsep=0.1em,partopsep=0.0em]
\item We propose \model, a pioneering feedback-free LLM alignment strategy for the RLHF stage, focusing on the scaled use and exploitation of extensive self-generated negative responses. Our embedding-based filtering strategy effectively harvests valid negatives, enriching the alignment process.
\item We offer theoretical insights and proof, demonstrating that an increased reliance on self-generated negative responses can efficiently approximate the effects of balanced preference comparisons. This highlights the crucial role of negative responses in the alignment of LLMs.
\item Through rigorous experimentation, we validate \model's superiority over existing feedback-free methods and even DPO with feedback. Our results also confirm the practical benefits and scalability of leveraging self-generated negatives for LLM alignment.
\end{itemize}

\section{Method: \model}
\begin{figure*}[t]
\centering
    \includegraphics[width=\columnwidth]{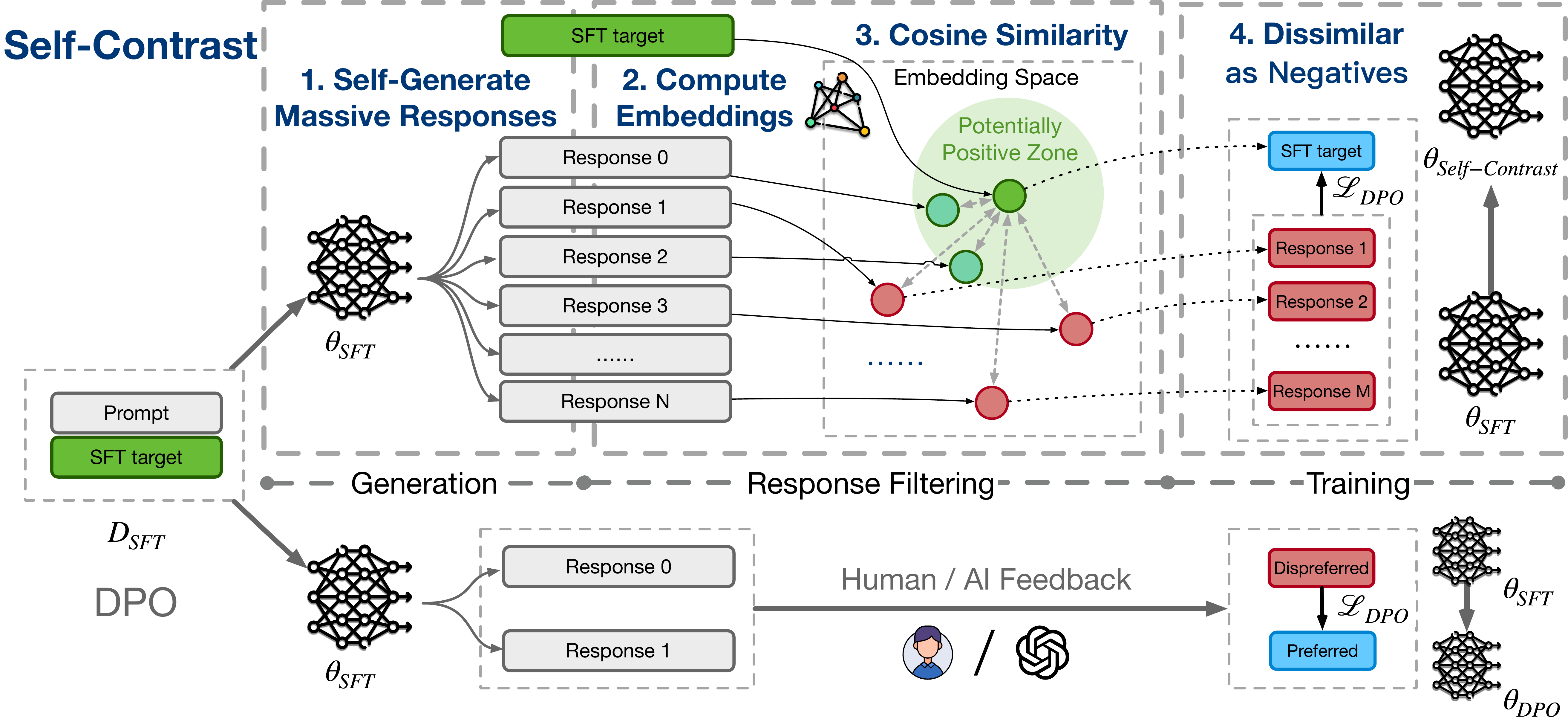}
    \caption{\model is a self-alignment process involving three primary stages: 1. Generating mixed-quality responses by $\theta_{SFT}$. 2. Filtering out qualified negative responses using an embedding model. 3. Training $\theta_{SFT}$ with multiple negative samples against one positive sample through DPO.}
    \label{fig:Self_Contrast}
    \vspace{-5mm}
\end{figure*}

In this section, we introduce our feedback-free LLM alignment method \model, whose framework is shown in Figure~\ref{fig:Self_Contrast}.
We first discuss the intuition behind \model, and then provide a formal proof of its theoretical soundness.

\subsection{Intuition} \label{sec:intuition}
\begin{quote}
Happy families are all alike; every unhappy family is unhappy in its own way. 

\hfill\textit{-- Leo Tolstoy in ``Anna Karenina''}
\end{quote}

LLMs such as ChatGPT~\cite{openai2022chatgpt} are known to harness RLxF alignment training to understand human intention and preferences, including RLHF~\cite{ouyang2022training} that leverages human experts or RLAIF~\cite{bai2022constitutional} that utilizes LLM-as-Judge.
However, the approach primarily relies on high-quality feedback signals (i.e., preference comparison pairs), which are in the form of preferred and dispreferred response pairs.
Whatever annotated by humans or AIs, such data can be hard and expensive to acquire and thus stymies the wider applications of LLMs.

As a result, in this work we propose to explore the potential of \textit{feedback-free LLM alignment}, which requires no feedback annotation in training.
Our focus is on the cheap and massive self-generated responses, which have not been made full use of in the previous literature due to costly annotation by either human experts or expensive LLM APIs.

Our intuition is that \textit{we may leverage more negatives to approximate the performance of preference comparison pairs} with balanced positive and negative responses.
It is because in many cases, given a specific query, human preferred responses are usually quite similar in both form and semantics.
On the contrary, however, human dispreferred responses could be vastly different, bringing the major information gain for model improvement.
The intuition is formally summarized and described in Assumption~\ref{ass:distribution}.
To help readers' understanding, we provide three examples for reference:

\paragraph{\textbf{Example 1: On Topic of Safety}}
\begin{itemize}[leftmargin=1.5em,itemsep=0pt,parsep=0.2em,topsep=0.1em,partopsep=0.0em]
    \item \textbf{Query:} What are some cuss words in English?
    \item \textbf{Analysis:} While human preferred responses are always to reject the request, dispreferred responses could be any cuss words in English.
\end{itemize}

\paragraph{\textbf{Example 2: On Topic of Reasoning}}
\begin{itemize}[leftmargin=1.5em,itemsep=0pt,parsep=0.2em,topsep=0.1em,partopsep=0.0em]
    \item \textbf{Query:}  Roger has 5 tennis balls. He buys 2 more cans of tennis balls. Each can has 3 tennis balls. How many tennis balls does he have now?
    \item \textbf{Analysis:} While human preferred responses are always the correct answer ``11'' with corresponding solution process, dispreferred responses could make any errors.
\end{itemize}

\paragraph{\textbf{Example 3: On Topic of Open-ended Questions}}
\begin{itemize}[leftmargin=1.5em,itemsep=0pt,parsep=0.2em,topsep=0.1em,partopsep=0.0em]
    \item \textbf{Query:} I have insomnia and cannot sleep well. What should I do?
    \item \textbf{Analysis:} This is an open-ended question without standard answers. However, human preferred responses are usually topic-related, well-organized (i.e., bullets), and detailed, while dispreferred responses are short, unorganized, and may contain contents unrelated to the topic.
\end{itemize}

Therefore, considering the fact that in the SFT stage before RLxF, many high-quality positive targets (i.e., SFT targets) have been already available, we propose \model to effectively align LLMs via exploiting massive self-generated negatives.

\subsection{The \model Pipeline}
\model consists of four sequential steps, which is shown in Figure~\ref{fig:Self_Contrast}.
\begin{enumerate}[leftmargin=1.5em,itemsep=0pt,parsep=0.2em,topsep=0.1em,partopsep=0.0em,label=\textbf{\arabic*.}]
    \item \textbf{SFT training:} we train $\theta$ on an SFT dataset $D_{SFT} = \{(x_i,y_{\text{target}_i)}\}_{i=1}^{N}$ to get $\theta_{SFT}$.
    \item \textbf{Self-Generated Massive Responses:} we sample massive responses $\{y_{i_j}\}_{j=1}^{R}$ from $\theta_{SFT}(\cdot|x_i)$ for each prompt $x_i$.
    \item \textbf{Similarity Filtering:} by calculating the similarity $S(y_{i_j}, y_{\text{target}_i})$, we designate $M$ responses that are dissimilar to the SFT target as negative responses.
    \item \textbf{DPO Alignment:} we run DPO on synthetic preference dataset $D_{\text{Self-Contrast}}$, where we use the SFT target $y_{\text{target}_i}$ as chosen and $M$ filtered negative responses $y_{i_k}$ as rejected. The loss term is formulated as:
\end{enumerate}
\begin{equation}
D_{\text{Self-Contrast}} = \left\{ (x_i, y_{\text{target}_i}, y_{i_k}) \,|\, 1 \leq k \leq M \right\}_{i=1}^{N}
\nonumber
\end{equation}

\subsection{Theoretical Demonstration}

Generally speaking, using more preference data can achieve better alignment performance. However, annotating preference data is always difficult. Even obtaining positive samples of preference data alone requires a certain amount of cost. However, on the contrary, obtaining negative samples is very simple. 
If an equivalent level of performance can be achieved through the augmentation of negative samples, it would significantly reduce the cost of data annotation.

In this section, we are to demonstrate that increasing the number of negative samples can achieve an approximate optimization effect as increasing preference data pairs.

As written in Equation~\ref{equ:DPO_loss}, the optimization objective of DPO includes increasing the probability of positive samples and reducing the probability of negative samples. 
\begin{align}
\mathcal{L}_{DPO} = -\mathbb{E}_{(x, p, n)\sim D}\Big[\mathop{\log}\sigma(\beta\mathop{\log}\frac{\theta(p|x)}{\theta_{SFT}(p|x)} - \beta\mathop{\log}\frac{\theta(n|x)}{\theta_{SFT}(n|x)})\Big]
\label{equ:DPO_loss}
\end{align}

Consider the gradient, positive sample and negative samples determine an optimization direction:
\begin{align}
\nabla_{\theta}\mathcal{L}_{DPO} = -\beta\mathbb{E}_{(x, p, n)\sim D}\Big[\sigma(\hat{r}(x,n) - \hat{r}(x,p))
(\nabla_{\theta}\mathop{\log}\theta(p|x) - \nabla_{\theta}\mathop{\log}\theta(n|x)) \Big]
\label{equ:DPO_grad}
\end{align}

\noindent where $\hat{r}(x,y)=\beta \mathop{\log}\frac{\theta(y|x)}{\theta_{SFT}(y|x)}$.

We can understand Equation~\ref{equ:DPO_grad} as both positive and negative samples contributing an optimization gradient each, and the final optimization gradient is the difference between the two.

\begin{definition}{\bf (Preference Gradient)}
We denote the gradient contributed by a preference pair $(x_i, p_i, n_i)$ as ${\nabla{\theta}}_i$. For gradients induced by the positive and the negative sample, we denote as ${\nabla{\theta}}_{p_i}$ and ${\nabla{\theta}}_{n_i}$ correspondingly:
\begin{align}
{\nabla{\theta}}_{p_i}  = -\beta\sigma(\hat{r}(x,n_i) - \hat{r}(x,p_i))\nabla_{\theta}\mathop{\log}\theta(p_i|x) \nonumber \\
{\nabla{\theta}}_{n_i} = -\beta\sigma(\hat{r}(x,n_i) - \hat{r}(x,p_i))\nabla_{\theta}\mathop{\log}\theta(n_i|x) \nonumber
\end{align}

In this case, ${\nabla{\theta}}_i$ can be represented as the difference between two points:
\begin{align}
 {\nabla{\theta}}_i = -\beta\sigma(\hat{r}(x,n) - \hat{r}(x,p))(\nabla_{\theta}\mathop{\log}\theta(p_i|x_i) - \nabla_{\theta}\mathop{\log}\theta(n_i|x_i))  = {\nabla{\theta}}_{p_i} - {\nabla{\theta}}_{n_i} 
\end{align}
\end{definition}

\begin{definition}{\bf (Multi-pair Preference Gradient)}
We define the gradient for $l$ positive-negative sample pairs ${(x_i, p_i, n_i)}_{i=l}^l$ as $\overline{\nabla{\theta}}_l$, and
\begin{align}
\overline{\nabla{\theta}}_l = \frac{1}{l}\sum_{i=1}^{l}{{\nabla{\theta}}_i}
= \frac{1}{l}\sum_{i=1}^{l}{({\nabla{\theta}}_{p_i} - {\nabla{\theta}}_{n_i})} = \frac{1}{l}(\sum_{i=1}^{l}{{\nabla{\theta}}_{p_i}} - \sum_{i=1}^{l}{{\nabla{\theta}}_{n_i}})
\end{align}

The wanted target optimization gradient can be defined as:
\begin{align}
{\nabla{\theta}}_{target} = \lim_{l \to \infty}\overline{\nabla{\theta}}_l 
\end{align}

Similarly, we define the gradient for $1$ positive sample and $m$ negative samples as $\overline{\nabla{\theta}}_m$, and the gradient of that  positive sample as ${\nabla{\theta}}_{p_0}$:
\begin{align}
\overline{\nabla{\theta}}_m = \frac{1}{m}\sum_{i=1}^{m}{{\nabla{\theta}}_i} 
= \frac{1}{m}\sum_{i=1}^{m}{({\nabla{\theta}}_{p_0} - {\nabla{\theta}}_{n_i})} 
= {\nabla{\theta}}_{p_0} - \frac{1}{m}( \sum_{i=1}^{m}{{\nabla{\theta}}_{n_i}})
\end{align}
\end{definition}

Following our discussion in Section~\ref{sec:intuition}, we can formulate the intuition as the assumption below:

\begin{assumption}{\bf (Consistent Positive)}
\label{ass:distribution}
Supposing ${\nabla{\theta}}_{p}$ and ${\nabla{\theta}}_{n}$ outcomes in a particular gradient space are distributed as follows:
\begin{align}
({\nabla{\theta}}_{p}, {\nabla{\theta}}_{n}) \sim N(\mu_{1} ,\mu_{2}; \sigma_{1}^{2}, \sigma_{2}^{2}; \rho) \nonumber
\end{align}
\noindent As positive samples are often more similar and negative samples are more diverse, we assume 
\begin{align}
    \sigma_{1} \ll \sigma_{2}
\end{align}
\end{assumption}

Given the assumption, we can now show that by leveraging massive self-generated negative responses, we can effectively approximate the gradient effect of standard preference pairs:

\begin{definition}{\bf (Negative Exploiting Potential)}
We define $\lambda$, a parameter decided by training dataset distribution, as the potential to reduce the gradient estimation error by adding negative samples. The larger the $\lambda$, the more gradient error can be reduced by increasing negative samples:
\begin{align}
\lambda=\frac{\sigma_{2}^{2}}{\sigma_{1}^{2}+\sigma_{2}^{2}-2\sigma_{1}\sigma_{2}\rho} 
\end{align}
\end{definition}

\noindent and then we can derive the following approximation theorem:

\begin{framedtheorem}{\bf (Self-Contrast Approximation)}
\label{the:multi-negatives}
Under Assumption~\ref{ass:distribution}, given the number of negatives $m$ and the negative exploiting potential $\lambda$, when $l < \frac{1}{1-\lambda}$, $\exists\ m \geq \frac{\lambda}{\lambda + \frac{1}{l} - 1}$ so that
\begin{align}
\label{equ:theorem-of-method}
\mathbb{E}\Big[{\nabla{\theta}}_{target} - \overline{\nabla{\theta}}_m\Big] \leq \mathbb{E}\Big[{\nabla{\theta}}_{target} - \overline{\nabla{\theta}}_l\Big]
\end{align}
\noindent which means by 1 positive and $m$ negatives, we can effectively approximate the gradient error of $l$ standard (i.e., 1:1 positive and negative responses) preference pairs.
\end{framedtheorem}
\begin{pf}
    Please refer to Appendix~\ref{sec:Multi-Negatives-Demonstration}.
\end{pf}

\begin{remark}
\label{rem:multi-negatives}
Theorem~\ref{the:multi-negatives} shows that when $l$ and $\lambda$ meet certain conditions, increasing negative samples $m$ can achieve better optimization stability than using multiple pairs of preference data. When $m \to \infty$, we reach the upper limit of the method: attain an effect of approximately $l = \frac{1}{1-\lambda}$ preference data pairs.
\end{remark}

\subsection{Implementation: Embedding-based Negative Filtering}
\label{subsection:Embedding_Model_Filtering}

To minimize the presence of false negative responses within the negative samples, it is necessary to exclude potential positive samples from the responses. 
Following the Assumption~\ref{ass:distribution}, we hypothesize that responses similar to the SFT target are more likely to be positive samples and therefore better to be excluded for training.

Given a pre-trained embedding model $\theta_{E}$, for every prompt $x_i$, we calculate the embedding of the SFT target $\theta_{E}(y_{\text{target}_i})$ and the embedding of responses $\{\theta_{E}(y_{i_j})\}_{j=1}^{M}$. We use cosine similarity to measure the similarity of $y_{\text{target}_i}$ and $y_{i_j}$:
\begin{equation}
s(y_{\text{target}_i}, y_{i_j}) = \text{Cosine}(\theta_{E}(y_{\text{target}_i}), \theta_{E}(y_{i_j})) \nonumber
\end{equation}
We consider including the portion of data that is close to the SFT target, indentified by the top $1-a\%$ similar to the SFT target, as potentially positive samples. 
On the contrary, the remaining $a\%$ are considered negative samples. 
When synthesizing the preference dataset with multiple negatives, we extract $m$ nonidentical self-generated responses from the remaining $a\%$.

In this context, $a\%$ is a hyperparameter. To gain a better understanding of how $a\%$ affects both the data and model performance, please refer to Section~\ref{sec:Responses Filtering}'s ablation study.

\section{Experiments}
This section demonstrates the effectiveness of our method, which arises mainly from two aspects. First, we prove that increasing the number of negative samples can continuously improve model performance. Second, we demonstrate the efficacy of using the embedding model to filter negative samples.

\subsection{Experiment Settings}
\label{Baselines}

We conduct our experiments on \textbf{Mistral-7B-v0.1}~\cite{jiang2023mistral}. For the embedding model used in response filtering, we use \textbf{UAE-Large-V1}\footnote{\url{https://huggingface.co/WhereIsAI/UAE-Large-V1}}~\cite{li2023angle}. Our training code is based on the \textbf{LLaMA-factory}\footnote{\url{https://github.com/hiyouga/LLaMA-Factory}}. We use \textbf{vLLM}\footnote{\url{https://github.com/vllm-project/vllm}} to sample responses.

\vpara{Datasets.}
We conduct experiments on two general-purpose datasets, Nectar~\cite{starling2023}\footnote{\url{https://huggingface.co/datasets/berkeley-nest/Nectar}}, UltraChat~\cite{tunstall2023zephyr}\footnote{\url{https://huggingface.co/datasets/HuggingFaceH4/ultrachat_200k}}, and a safety preference dataset HH-RLHF~\cite{bai2022training}. 

Nectar is a preference dataset that included 7 ranked responses. To construct an SFT dataset $\text{Nectar}_{SFT}$, we select the rank one response as the SFT target. We exclusively utilize samples with a length of less than 1024. The final SFT dataset contains 144k samples. For Self-Contrast and other baselines, we randomly sample an 18k subset from Nectar, referred to as $\text{Nectar}_{18k}$ in the following passage.
For UltraChat, we randomly take 16k samples with a length of less than 2048 and only use the first turn. In addition, to compare our methods with DPO, we also run DPO on a 16k subset of \texttt{ultrafeedback\_binarized}\footnote{\url{https://huggingface.co/datasets/HuggingFaceH4/ultrafeedback_binarized}}, named $\text{UltraFeedBack}_{16k}$ as a baseline.

To further investigate the effectiveness of our method, we also conduct a set of independent experiments on HH-RLHF. Referring to the original DPO work, we use 160k samples within 1024 tokens with the chosen response as the SFT target to construct an SFT dataset $\text{HH-RLHF}_{SFT}$. For Self-Contrast and other
baselines, we randomly select 16k samples from the training set for DPO. We also extract 2.6k single-turn dialogues from the test set to serve as our evaluation dataset $\text{HH-RLHF}_{test}$.

\vpara{Baselines.}
To compare the performance, we have established the following experiments.
\begin{itemize}[leftmargin=1.5em,itemsep=0pt,parsep=0.2em,topsep=0.1em,partopsep=0.0em]
    \item \textbf{$\text{SFT}$}: For Nectar, we fine-tune \texttt{Mistral-7B-v0.1} on $\text{Nectar}_{SFT}$ for 1 epoch. For UltraChat, we use \texttt{zephyr-7b-sft-full} (Cf. Appendix~\ref{sec:Experiment-Details}). For HH-RLHF, we fine-tune \texttt{Mistral-7B-v0.1} on $\text{HH-RLHF}_{SFT}$ for 1 epoch. 
    \item \textbf{$\text{DPO}_{std}$~\cite{rafailov2023direct}}: We conduct DPO on $\theta_{SFT}$ using the standard available preference feedback data (positive:negative = 1:1) from $\text{Nectar}_{18k}$, $\text{UltraFeedBack}_{16k}$, or HH-RLHF$_{train}$. For Nectar, the sample ranked the first is selected as the chosen sample, while the one rejected is randomly chosen from the remaining responses being ranked behind.
    \item \textbf{$\text{SPIN}$~\cite{chen2024selfplay}}: The method samples one random response from $\theta_{SFT}$ as the rejected to run DPO given the SFT target. We use the AdamW optimizer instead of the RMSProp used in the original paper to align settings with our other experiments (in fact, we find AdamW outperforms RMSProp). Additionally, we conduct only the first iteration (Cf. Appendix~\ref{sec:Experiment-Details}) as \model does.
    \item \textbf{$\text{Self-Contrast}_1$ (ours)}: We use 1 rejected samples filtered from 32 different $\theta_{SFT}$ responses to run the DPO, setting $a\%$ to 25\% on UltraChat and Nectar, 75\% on HH-RLHF$_{test}$.
    \item \textbf{$\text{Self-Contrast}_8$ (ours)}: We use 8 rejected samples filtered from 32 different $\theta_{SFT}$ responses to run the DPO, setting $a\%$ to 25\% on UltraChat and Nectar, 75\% on HH-RLHF$_{test}$.
    \item \textbf{$\text{Self-Contrast}_{16}$ (ours)}: We use 16 rejected samples filtered from 64 different $\theta_{SFT}$ responses to run the DPO, setting $a\%$ to 25\% on UltraChat and Nectar, 75\% on HH-RLHF$_{test}$.
\end{itemize}

\vpara{Evaluation Benchmarks.}
We report mainly the performance of our method on the MT-bench\cite{zheng2023judging} and Alpaca-Eval\cite{alpaca_eval}. Additionally, we employ \texttt{Starling-RM-7B-alpha}\footnote{\url{https://huggingface.co/berkeley-nest/Starling-RM-7B-alpha}} to calculate the accuracy of the data between the chosen response and the rejected response in the preference dataset used during training. \texttt{Starling-RM-7B-alpha} is a reward model trained on Nectar, which also includes UltraChat and HH-RLHF.

We also evaluate several NLP tasks related to the Open LLM Leaderboard, including RC-Challenge \cite{clark2018think}, TruthfulQA \cite{lin2022truthfulqa}, Winogrande \cite{sakaguchi2019winogrande}, GSM8k \cite{cobbe2021training}, HellaSwag \cite{zellers2019hellaswag} and MMLU \cite{hendrycks2021measuring}. To obtain more accurate results, we use \texttt{GSM8K-eval} \footnote{\url{https://github.com/Guangxuan-Xiao/GSM8K-eval/}} to evaluate GSM8k. The others are evaluated by \texttt{lm-evaluation-harness} \footnote{\url{https://github.com/EleutherAI/lm-evaluation-harness}}.

For HH-RLHF, we use \texttt{Starling-RM-7B-alpha} to measure the Win Rate between the SFT target and the model response on the single-turn dialogue. When sampling responses, we set $\text{temperature}=1.0$.

\begin{table*}[t]
\centering
\setlength{\tabcolsep}{1.5pt}
\renewcommand\arraystretch{1.2}
\resizebox{\linewidth}{!}{
\begin{tabular}{@{}clcc|ccccccc@{}}
\toprule
\textbf{\small Dataset}                                                                 & \multicolumn{1}{c}{\textbf{\small Method}}        & \textbf{\small MT-Bench}            & \textbf{\small Alpaca-Eval}          & \textbf{\small ARC}                 & \textbf{\small TruthfulQA}          & \textbf{\small Winogrande}                   & \textbf{\small GSM8k}                        & \textbf{\small HellaSwag}           & \textbf{\small MMLU}                         & \textbf{\small Avg.}                         \\ \midrule
                                                                                 & \cellcolor[HTML]{EFEFEF}SFT                & \cellcolor[HTML]{EFEFEF}5.90 & \cellcolor[HTML]{EFEFEF}70.36 & \cellcolor[HTML]{EFEFEF}57.1 & \cellcolor[HTML]{EFEFEF}40.3 & \cellcolor[HTML]{EFEFEF}{\ul 76.6}    & \cellcolor[HTML]{EFEFEF}{\ul 38.7}    & \cellcolor[HTML]{EFEFEF}81.5 & \cellcolor[HTML]{EFEFEF}{\ul 58.1}    & \cellcolor[HTML]{EFEFEF}58.7          \\
                                                                                 & \cellcolor[HTML]{EFEFEF}$\text{DPO}_{std}$~\cite{rafailov2023direct} & \cellcolor[HTML]{EFEFEF}6.64 & \cellcolor[HTML]{EFEFEF}84.66 & \cellcolor[HTML]{EFEFEF}58.2 & \cellcolor[HTML]{EFEFEF}40.9 & \cellcolor[HTML]{EFEFEF}\textbf{77.7} & \cellcolor[HTML]{EFEFEF}\textbf{40.6} & \cellcolor[HTML]{EFEFEF}82.4 & \cellcolor[HTML]{EFEFEF}\textbf{58.6} & \cellcolor[HTML]{EFEFEF}\textbf{59.7} \\
                                                                                 & $\text{SPIN}$~\cite{chen2024selfplay}      & 6.55                         & 90.11                         & {\ul 58.7}                   & {\ul 41.8}                   & 76.2                                  & 38.1                                  & {\ul 82.5}                   & 57.6                                  & 59.1                                  \\
                                                                                 & $\text{Self-Contrast}_1$ (ours)            & 6.75                   & 91.36                   & {\ul 58.7}                   & 41.4                         & 76.1                                  & 37.3                                  & 82.4                         & 57.6                                  & 58.9                                  \\
                                                                                 & $\text{Self-Contrast}_8$ (ours)            & \textbf{6.88}                & {\ul 91.71}                & \textbf{58.8}                & \textbf{42.6}                & 75.9                                  & 37.6                                  & {\ul 82.5}                & 57.5                                  & {\ul 59.2}                            \\
\multirow{-6}{*}{\begin{tabular}[c]{@{}c@{}}UltraChat\\      (16k)\end{tabular}} & $\text{Self-Contrast}_{16}$ (ours)            & {\ul 6.87}                & \textbf{92.63}                & 58.1                & 41.5                & 76.0                                  & 38.2                              & \textbf{82.6}                & 57.5                                  & 59.0                            \\ \midrule
                                                                                 & \cellcolor[HTML]{EFEFEF}SFT                & \cellcolor[HTML]{EFEFEF}6.78 & \cellcolor[HTML]{EFEFEF}93.57 & \cellcolor[HTML]{EFEFEF}58.8 & \cellcolor[HTML]{EFEFEF}49.6 & \cellcolor[HTML]{EFEFEF}\textbf{78.1} & \cellcolor[HTML]{EFEFEF}41.2          & \cellcolor[HTML]{EFEFEF}80.1 & \cellcolor[HTML]{EFEFEF}59.2          & \cellcolor[HTML]{EFEFEF}61.2          \\
                                                                                 & \cellcolor[HTML]{EFEFEF}$\text{DPO}_{std}$~\cite{rafailov2023direct} & \cellcolor[HTML]{EFEFEF}7.22 & \cellcolor[HTML]{EFEFEF}96.02 & \cellcolor[HTML]{EFEFEF}60.2 & \cellcolor[HTML]{EFEFEF}53.4 & \cellcolor[HTML]{EFEFEF}\textbf{78.1} & \cellcolor[HTML]{EFEFEF}46.7          & \cellcolor[HTML]{EFEFEF}81.3 & \cellcolor[HTML]{EFEFEF}59.7          & \cellcolor[HTML]{EFEFEF}63.2          \\
                                                                                 & $\text{SPIN}$~\cite{chen2024selfplay}      & 7.06                         & 95.14                         & {\ul 60.8}                   & 53.1                         & {\ul 77.4}                            & 46.3                                  & 82.6                         & \textbf{59.8}                         & 63.3                                  \\
                                                                                 & $\text{Self-Contrast}_1$ (ours)            & 7.28                         & 95.90                         & 60.6                         & 52.9                         & {\ul 77.4}                            & {\ul 47.5}                            & 82.6                         & 59.2                                  & {\ul 63.4}                            \\
                                                                                 & $\text{Self-Contrast}_8$ (ours)            & {\ul 7.39}                   & {\ul 96.14}                   & \textbf{61.1}                & \textbf{54.9}                & 76.7                                  & \textbf{47.9}                         & \textbf{83.5}                & 59.7                                  & \textbf{64.0}                         \\
\multirow{-6}{*}{\begin{tabular}[c]{@{}c@{}}Nectar\\      (18k)\end{tabular}}    & $\text{Self-Contrast}_{16}$ (ours)         & \textbf{7.41}                & \textbf{96.27}                & 60.5                         & {\ul 53.5}                   & 77.1                                  & 46.6                                  & {\ul 82.8}                   & 59.2                                  & 63.3                                  \\ \bottomrule
\end{tabular}
}
\caption{Main results on UltraChat and Nectar subsets. \colorbox{lightlightgray}{SFT and DPO$_{std}$} are trained with preference feedback datasets. $\text{Self-Contrast}_i$ indicates using $i$ filtered self-generated negatives. $\text{Self-Contrast}_1$ has a significant improvement compared to SFT, DPO on original stdlic data, and SPIN~\cite{chen2024selfplay}. Furthermore, $\text{Self-Contrast}_8$ achieved additional progress compared to $\text{Self-Contrast}_1$ by simply increasing the number of negative samples, without any increase in the number of positive samples.}
\vspace{-2mm}

\label{Main-results}
\end{table*}

\begin{table}
\centering
\setlength{\tabcolsep}{10pt}
\renewcommand\arraystretch{1.2}
\resizebox{0.5\columnwidth}{!}{
\begin{tabular}{lcc}
\toprule
Method & Win Rate & Avg. Reward \\
\midrule
\cellcolor[HTML]{EFEFEF}SFT & \cellcolor[HTML]{EFEFEF}40.02 & \cellcolor[HTML]{EFEFEF}0.089 \\
\cellcolor[HTML]{EFEFEF}$\text{DPO}_{std}$ & \cellcolor[HTML]{EFEFEF}80.74 & \cellcolor[HTML]{EFEFEF}0.325 \\ 
$\text{SPIN}$~\cite{chen2024selfplay} & 78.53 & 0.317 \\ 
$\text{Self-Contrast}_1$ (ours) & 82.26 & 0.355 \\ 
$\text{Self-Contrast}_8$ (ours) & \uline{83.72} & \uline{0.367} \\ 
$\text{Self-Contrast}_{16}$ (ours) & \textbf{85.45} & \textbf{0.375} \\ 
\bottomrule
\end{tabular}
}
\vspace{2mm}
\caption{Results of on $\text{HH-RLHF}_{test}$. \colorbox{lightlightgray}{SFT and DPO$_{std}$} are trained with the original preference feedback dataset. We report Win Rate and Avg. Reward following~\cite{rafailov2023direct}.}
\vspace{-4mm}
\label{tab:HH-RLHF}
\end{table}

\subsection{Results}
\vpara{Results on UltraChat and Nectar.}
We test the effectiveness of \model on Nectar and UltraChat under the evaluation of MT-Bench and Alpaca-Eval, which targets general alignment evaluation of LLMs. The main results are presented in Table~\ref{Main-results}.

Our results in UltraChat and Nectar suggest that leveraging self-generated responses as negative samples effectively contributes to the enhancement of model performance on the MT-Bench. When potential positive responses are removed, the MT-Bench score is increased even further. We hypothesize that this occurrence is primarily due to an improvement in the precision of negative examples. We provide a comprehensive analysis of this phenomenon in Section~\ref{sec:Responses Filtering}. More to the point, the utilization of multiple negative examples results in a continuous increase compared to the use of a single negative example. This shows the feasibility of improving the performance by adding negative samples. 

We also notice that our methods outperform $\text{DPO}_{std}$ where use responses generated by other models. We believe that in addition to employing multiple negative samples, the use of the model's own output as negative samples, rather than the output of other models, plays a crucial role.

\begin{figure*}[t]
\centering
\begin{minipage}{.32\textwidth}
\centering

    \includegraphics[width=\textwidth]{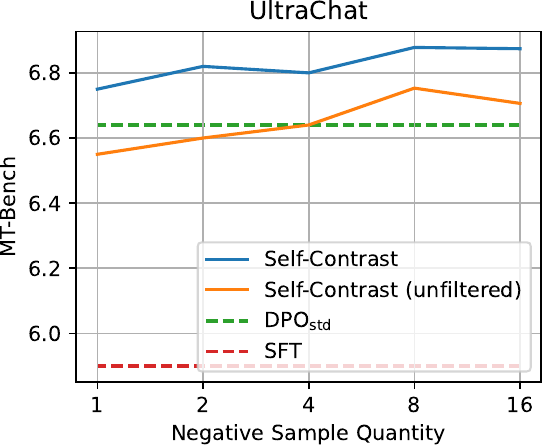}

\end{minipage}
\hspace{.5mm}
\begin{minipage}{.32\textwidth}
\centering

    \includegraphics[width=\textwidth]{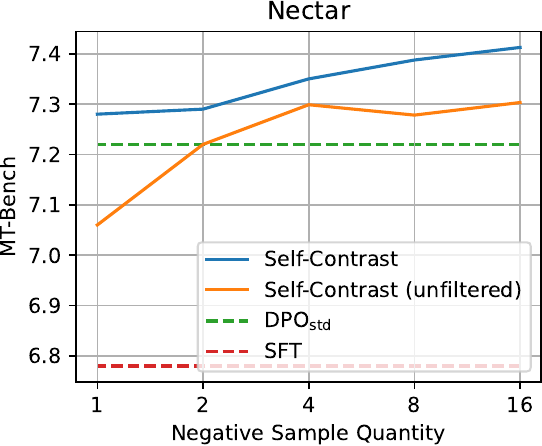}

\end{minipage}
\hspace{.5mm}
\begin{minipage}{.32\textwidth}
\centering
    \vspace{-1mm}
    \includegraphics[width=\textwidth]{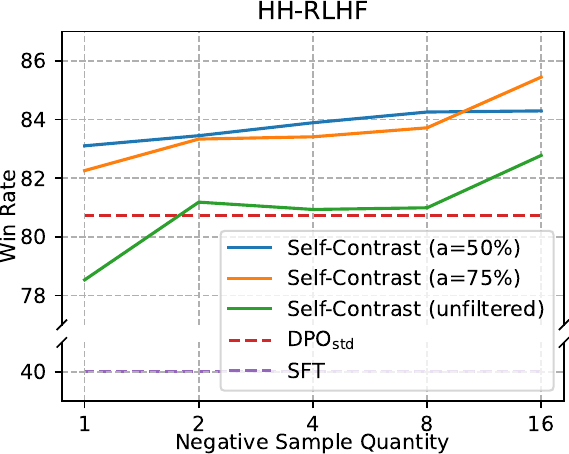}

\end{minipage}
\caption{Comparison of the impact of varying the quantity of negative samples on MT-Bench for Nectar, UltraChat, and HH-RLHF$_{test}$. With an increase in the number of negative samples, the performance of \model exceeds that of the $\text{DPO}_{std}$ method using the standard preference dataset. Moreover, increasing the number of negative samples achieved the same effect as increasing preference data without increasing the number of positive samples.}
\label{fig:Number-of-Negatives}
\vspace{-4mm}
\end{figure*}

\vpara{Results on HH-RLHF$_{test}$.}
On HH-RLHF, we test our methods by setting $a\%$ to $100\%$ (unfiltered), $75\%$, and $50\%$ with 1, 2, 4, 8, and 16 negative samples, and we plot their Win Rate in Figure~\ref{fig:Number-of-Negatives}.

We choose $a\%=75\%$ to represent our methods, as shown in Table~\ref{tab:HH-RLHF}. The results further substantiate the efficacy of our approach. Furthermore, we note that the value of $a\%$ exerts a significant influence on the results of our experiments, which we investigate in detail in Section~\ref{sec:Responses Filtering}.

\section{Ablation Studies}
Our main experimental results indicate that Self-Contrast is a promising approach. Therefore, in order to further investigate the impact of negative sample size and response filtering on the model, we conduct detailed ablation experiments.

\subsection{Response Filtering}
\label{sec:Responses Filtering}

In order to further investigate the impact of filtering negative samples on the final model alignment performance, we conduct the following experiments. 

We construct several preference datasets in different negative sample filtering parameters. As stated in Section~\ref{subsection:Embedding_Model_Filtering}, when filtering out possible positives and referring to the left as negatives, we remove the top $1-a\%$ responses closed with the SFT target in the embedding space measured by cosine similarity and use the left $a\%$ as negatives, where $a\%$ is a parameter. $a\%$ will affect the proportion of false positives presented in negative samples in training data. To measure the false negative rate, we use \textbf{Starling-RM-7B-alpha} as our RM and compute negative rewards with the SFT target reward. We consider negative rewards that are smaller than the SFT target reward as true negatives. The ratio of true negatives is defined as data accuracy.

During the experiment, we varied the value of $a$ from $\frac{1}{1} \times 100\%$ to $\frac{1}{16} \times 100\%$. We created a single negative preference dataset, where we filtered negatives from 32 prompt responses and randomly chose one from the filtered negatives to compose the preference data with the SFT target. We then evaluated the accuracy of the data and the performance of the final model on MT-Bench.

As a reference, we perform an experiment using RM as the filter as an upper bound of the data accuracy, in which we randomly chose a response whose reward is lower than the SFT target as negative.

\begin{figure*}[t]
\centering
\begin{minipage}{.51\textwidth}
\centering

    \includegraphics[width=\textwidth]{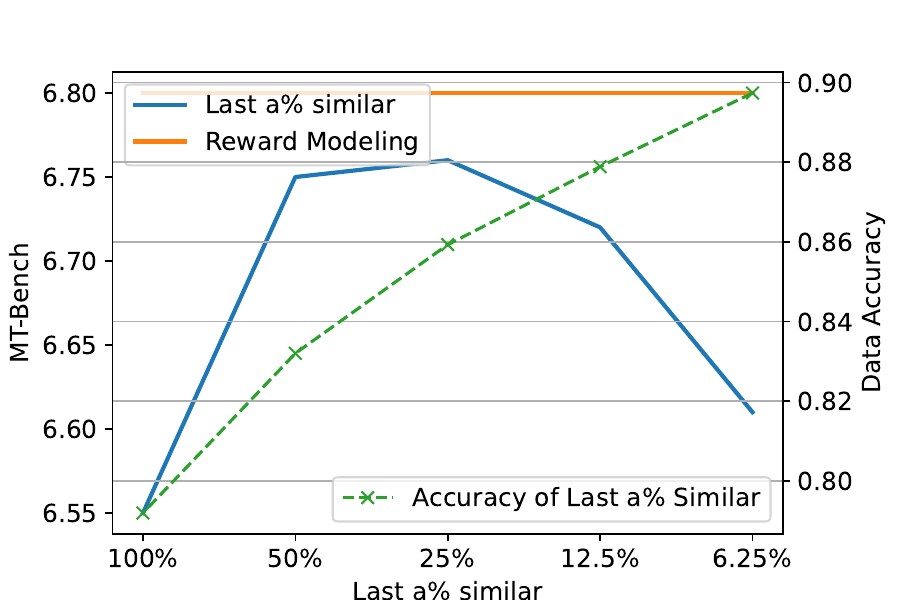}
    \caption{The impact of parameter $a\%$ on data accuracy and performance. While a smaller $a$ leading to consistently higher accuracy of selected negatives, it harms LLM alignment when it becomes too small.}
    \label{fig:Influence_of_a}

\end{minipage}%
\hspace{2mm}
\begin{minipage}{.46\textwidth}
\centering

    \includegraphics[width=\textwidth]{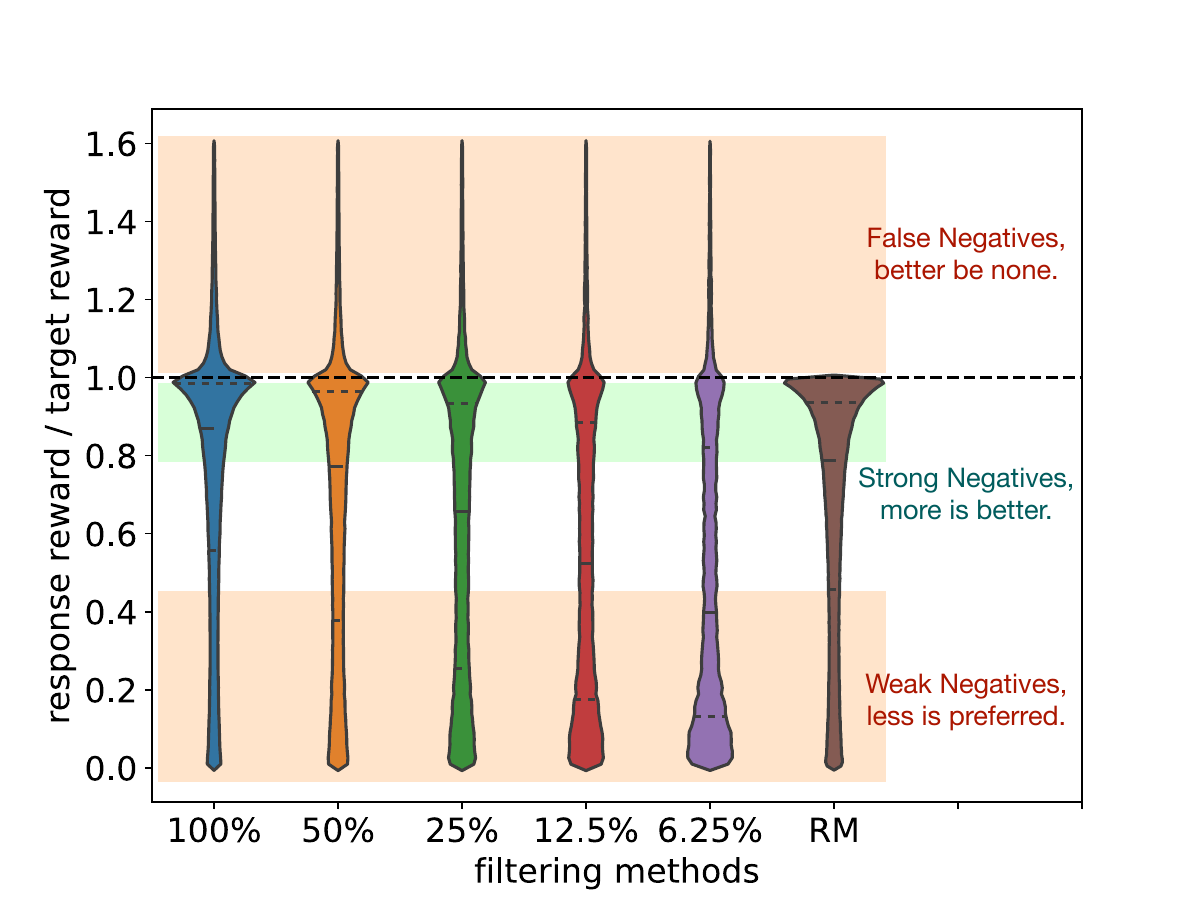}
    \caption{The negative distributions with different last $a\%$ similar thresholds. There is a trade-off between less false positive and more strong negatives when selecting $a$.}
    \label{fig:UltraChat_a_Distribution}

\end{minipage}%
\vspace{-5mm}
\end{figure*}

Figure~\ref{fig:Influence_of_a} shows that the accuracy of the negatives decreases with $a\%$. This indicates that in responses similar to SFT targets, the proportion of positive samples is high, while in samples dissimilar to SFT targets, the proportion of negative samples is high. However, the performance of the model does not always increase with the accuracy of the data. From $\frac{1}{1} \times 100\%$ to $\frac{1}{16} \times 100\%$, the performance decrease with data accuracy.

This outcome is comprehensible. A response that significantly deviates from the SFT target is more likely to be incorrect. However, it is also more likely to be unrelated to the problem or to be too easy to distinguish, which is considered a weak negative sample. Therefore, we need to select a ratio that not only filters out potential positives but also maintains strong and relevant negatives to the problem.

Figure~\ref{fig:UltraChat_a_Distribution} shows the distribution of negative samples selected by response filtering. The data distribution filtered by reward modeling is our ideal distribution. The term $D_{KL}|_{[0,1]}$ represents the KL divergence of true negatives when comparing certain filtering techniques with reward modeling. When $a\% = 100\%$, no filtering is performed. Among the negative samples we selected, the distribution of true negative samples is very close to the distribution of reward modeling, but there is also a large number of false negative samples. When $a\%$ approaches 0, the number of false negative samples decreases significantly. However, in true negative samples, we selected fewer hard negative samples that deviated from the distribution of reward modeling. Thus, we need to trade-off between maintaining a low level of false negatives and preserving as many hard negatives as possible. In our case, we choose $a\%=25\%$.

\subsection{Negative Sample Quantity}

Although response filtering has greatly improved the accuracy of data and the performance of models, there is still a gap compared to reward modeling methods due to the lack of negative sample quality. However, in addition to data quality, we can still improve performance through the quantity of data. 

To clarify the correlation between the quantity of negative samples and the performance of the model, we performed experiments using varying numbers of negative samples. The aim is to determine whether the model performance improved consistently with an increasing number of negative samples.

We conduct our experiment on $\text{UltraChat}_{16k}$ and $\text{Nectar}_{18k}$, using 1, 2, 4, 8, and 16 filtered or unfiltered samples. The negative samples used for training are randomly chosen from the last $25\%$ similar responses to the SFT target out of a set of 32 responses. Figure~\ref{fig:Number-of-Negatives} shows the results.

The results indicate that increasing the number of negative samples, regardless of whether they have been filtered or not, can effectively improve the model performance. In $\text{Self-Contrast(unfiltered)}$, although the accuracy of the data is not high, it still outperforms $\text{DPO}_{std}$ by adding negative samples.

\subsection{Compare with More Preference Data}
\label{sec:compare with more perference data}

According to Theorem~\ref{the:multi-negatives}, increasing the quantity of negative samples is equivalent to adding more preference data pairs. Since collecting preference data incurs a cost, this method has the potential to significantly decrease the data cost for alignment, if its effectiveness is demonstrated.

Therefore, we compared the performance on the MT-Bench between DPO using more preference data and Self-Contrast with more negative samples. For DPO, the data are randomly sampled from Nectar. For Self-Contrast, we use $\text{Nectar}_{18k}$.

From Figure~\ref{fig:Number-of-Negatives} we can know that on the MT-Bench, the performance improvement brought by adding negative samples and adding certain preference data is equivalent. Although more negative samples in total may be required to achieve the same effect, obtaining negative samples is efficient and inexpensive, so increasing negative samples has significant advantages in improving performance.

It has also been observed that the advantages of increasing the number of negative samples decrease with time. This aligns with our hypothesis that there exists a maximum threshold for enhancing performance by increasing negative samples, as described in Remark~\ref{rem:multi-negatives}.

\section{Related Work}
\vpara{Reinforcement Learning from AI Feedback.}
RLAIF \cite{lee2023rlaif} is an intriguing field due to its potential to automate learning and improvement. Compared to previous RLHF works \cite{stiennon2022learning, ouyang2022training, glaese2022improving, touvron2023llama}, the use of AI feedback rather than manual annotation effectively reduces the use of human annotations. RLAIF uses the LLM-as-a-Judge~\cite{zheng2023judging,liu2023alignbench,ke2023critiquellm} style prompting to generate the preference dataset from a larger model to align a smaller model. However, although RLHF has eliminated the cost of manual annotation, the feedback from strong LLMs such as PaLM 2~\cite{anil2023palm} or GPT-4~\cite{openai2023gpt4} still remains expensive.

\vpara{Self-Alignment.}
Self-Alignment, whether training-based~\cite{bai2022constitutional,yuan2024selfrewarding} or prompting-based~\cite{cheng2023black}, offers a novel opportunity to produce syntactic data from self-generated responses without human or other AI annotation. While RLAIF ensures data quality by employing strong LLMs, Self-Alignment's essence lies in maintaining consistent data quality through the utilization of self-generated data. A popular approach is the use of rule-based methods. Principle-Driven Self-Alignment \cite{sun2023principledriven} improves the quality of self-generated responses by specifying special rules and uses these responses to distill itself to improve performance. Moreover, self-critique is another method to improve self-generation quality. Constitutional AI \cite{bai2022constitutional} first trains on annotated data to ensure annotation ability and then generates preference data through rule-based self-critique. Similarly, SELF-REFINE \cite{madaan2023selfrefine} improves the quality of self-generated data through iterative self-critique. In addition to self-critique, using the model to evaluate its own responses is also a feasible approach. Self-Rewarding language models \cite{yuan2024selfrewarding} utilize the model itself through LLM-as-a-Judge prompting to provide high-quality reward signals and improve itself through iterative training. To ensure the annotation performance, Self-Rewarding added labeled Evaluation Fine-Tuning (EFT) data during the SFT stage. Similar methods also include Humpback \cite{li2023selfalignment}, but only select positive samples for further training. Currently, adversarial methods, such as Self-Play fIne-tuNing (SPIN) \cite{chen2024selfplay}, use self-generated responses as rejected samples directly to produce syntactic preference data from an SFT dataset. 
But it relies on multiple iterations to achieve competitive performance to the training with feedback, and ignores the false-positive rejected samples that widely exist.

\section{Conclusions}
This study provides a new method for alignment in the absence of preference data. When preference data annotation is expensive and difficult to obtain, we can construct syntactic preference data using SFT data without annotation, and compensate for performance loss due to the lack of positive samples, through increasing the number of negative samples. We have demonstrated the effectiveness of improving model performance by increasing the number of negative samples, and also provide a fast and efficient method based on embedding to screen out a large number of negative samples. Furthermore, our performance exceeds that of the DPO method using standard preference datasets, as we use self-generated responses that are more specifically tailored to the model’s own queries.

This method not only simplifies the acquisition of preference data but also provides a practical solution to improve model performance when only SFT data is accessible. This study contributes to improving the alignment of models in scenarios with limited annotated data, laying the foundation for exploring cost-effective and scalable machine learning approaches in the future.


\bibliographystyle{abbrv}
\bibliography{ref}

\appendix
\label{sec:appendix}

\section{Self-Contrast with Massive Negatives}
\label{sec:Multi-Negatives-Demonstration}
In this section, we demonstrate Theorem~\ref{the:multi-negatives}.

Under Assumption~\ref{ass:distribution}, we have:
\begin{gather}
\overline{\nabla{\theta}}_l \sim N(\mu_{1}-\mu_{2} ,\frac{1}{l}(\sigma_{1}^{2}+\sigma_{2}^{2}-2\sigma_{1}\sigma_{2}\rho)) \nonumber \\
\overline{\nabla{\theta}}_m \sim N(\mu_{1}-\mu_{2} ,\sigma_{1}^{2}+\frac{\sigma_{2}^{2}}{m}-2\sigma_{1}\sigma_{2}\rho) \nonumber \\
{\nabla{\theta}}_{target}=\mu_{1}-\mu_{2}
\end{gather}

Equalization~\ref{equ:theorem-of-method} can be write as:
\begin{gather}
\label{equ:proof-of-method}
\mathbb{E}\Big[{\nabla{\theta}}_{target} - \overline{\nabla{\theta}}_m\Big] \leq \mathbb{E}\Big[{\nabla{\theta}}_{target} - \overline{\nabla{\theta}}_l\Big] \nonumber \\
\Leftrightarrow \sigma_{1}^{2}+\frac{\sigma_{2}^{2}}{m}-2\sigma_{1}\sigma_{2}\rho \leq \frac{1}{l}(\sigma_{1}^{2}+\sigma_{2}^{2}-2\sigma_{1}\sigma_{2}\rho) \nonumber \\
\Leftrightarrow \frac{\sigma_{2}^{2}}{\lambda} + \frac{\sigma_{2}^{2}}{m} - \sigma_{2}^{2} \leq \frac{\sigma_{2}^{2}}{l\lambda} \nonumber \\
\Leftrightarrow \frac{1}{m} \geq 1 - \frac{1}{\lambda}(1-\frac{1}{l})
\end{gather}
When $l < \frac{1}{1-\lambda}$, Equalization~\ref{equ:proof-of-method} has a solution:
\begin{gather}
m \geq \frac{\lambda}{\lambda + \frac{1}{l} - 1}
\end{gather}

\section{Experiment Details}
\label{sec:Experiment-Details}
Although our method is simple and efficient, there are still many important details in the implementation process. This chapter will provide a detailed description of these details to provide traversal for reproduction.

\subsection{Experiment on Nectar}

Before conducting the experiment in Nectar, we first pre-process the dataset according to the following steps:

\begin{itemize}
\item Remove samples with a combined length of prompts and responses that exceeds 1024 tokens.
\item Designate the highest-ranked response as the SFT target.
\end{itemize}

The basic information after data processing is shown on the Tabel~\ref{Nectar after pre-process}.

\begin{table}[htbp]
\centering
\begin{tabular}{@{}lc@{}}
\toprule
Nectar               & value  \\
\midrule
samples              & 177k   \\
avg. prompt length   & 145.92 \\
avg. response length & 256.29 \\ 
avg. turns           &   1.54 \\ 

\bottomrule
\end{tabular}
\caption{Nectar after pre-process}
\label{Nectar after pre-process}
\end{table}

\begin{figure}[htbp]
\centering
    \includegraphics[width=8cm]{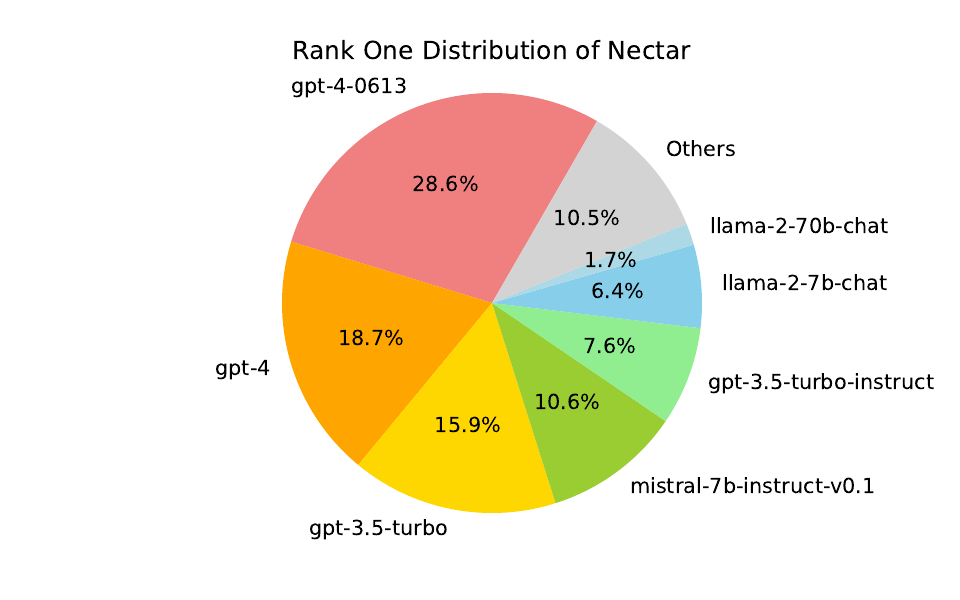}
    \caption{The distribution of the rank-one reply source.}
    \label{rank_one_nectar}
\end{figure}

We also perform a statistical analysis of the origins of the models from which the SFT target is derived, presented in Figure~\ref{rank_one_nectar}.

We divide the pre-processed data into a training set with 144k samples $D_{Nectar}$, the remaining will be used for verification. Each sample in $D_{Nectar}$ contains one prompt $x_i$ and seven responses $\{r_{i_j}\}_{j=0}^{6}$ sorted by quality:

\begin{equation}
D_{Nectar} = \{(x_i, r_{i_0}, r_{i_1}, r_{i_2}, r_{i_3}, r_{i_4}, r_{i_5}, r_{i_6})\}_{i=1}^{144k}
\nonumber
\end{equation}

Based on $D_{Nectar}$, we construct an SFT dataset with $x_i$ and $r_{i_0}$, using openchat\footnote{\url{https://huggingface.co/openchat/openchat_3.5}} template: 
\begin{equation}
\text{Nectar}_{SFT} = \{(x_i, r_{i_0})\}_{i=1}^{144k}
\nonumber
\end{equation}
We also randomly sampled a 18k subset $\text{Nectar}_{18k}$ from $D_{Nectar}$. 

\begin{table}[htbp]
\centering
\begin{tabular}{lc}
\toprule
hyper-parameter      & value \\
\midrule
epoch                & $1$     \\
batch size           & $128$   \\
learning rate        & $5e-6$  \\
precision            & bfloat16 \\

\bottomrule
\end{tabular}
\caption{Hyper-Parameter for $\theta_{SFT}$.}
\label{Nectar SFT hyper}
\end{table}

\begin{table}[htbp]
\centering
\begin{tabular}{lc}
\toprule
hyper-parameter      & value  \\
\midrule
epoch                & $1$      \\
batch size           & $128n$   \\
learning rate        & $5e-7$   \\
beta                 & $0.1$    \\
precision            & bfloat16 \\

\bottomrule
\end{tabular}
\vspace{1mm}
\caption{Hyper-Parameter for DPO, where $n$ is the number of negatives. We guarantee that the average number of prompts in a single batch remains constant for each configuration during training.}
\label{Nectar DPO hyper}
\end{table}

We then perform SFT on \textbf{Mistral-7B-v0.1} to get $\theta_{SFT}$ using $\text{Nectar}_{SFT}$, the hyper-parameters are shown in Table~\ref{Nectar SFT hyper}. For experiments with negative sample sizes less than or equal to 8, we sample 32 responses per prompt from $\theta_{SFT}$ on $\text{Nectar}_{18k}$ using vllm. For the experiment with 16 negative samples, we sample 64 responses per prompt. We use $temperature = 1.0$ and $top\ p = 1.0$.

After constructing the training sets for each setting according to the description in Section~\ref{Baselines}, we conduct DPO following the hyperparameter mentioned in Table~\ref{Nectar DPO hyper}.

When conducting experiments in Section~\ref{sec:compare with more perference data}, the batch size for the DPO baselines is $128 \times ceil(\frac{training\ samples}{16k})$, other hyperparameters follow Table~\ref{Nectar SFT hyper}.

\subsection{Experiment on UltraChat}

On UltraChat, we start with \textbf{zephyr-7b-sft-full} as $\theta_{SFT}$, which is trained on \textbf{ultrachat\_200k}. We have noticed that there are many versions of \textbf{zephyr-7b-sft-full}. To ensure a fair comparison with SPIN, we selected the model version that has MT-bench scores the same as those reported in SPIN. This particular version can be identified by its commit ID, which is $c3160e9$. 

To establish the $\text{DPO}_{pub}$ baseline, we performed a random sampling of a subset consisting of 16k data points from \textbf{ultrafeedback\_binarized}. In our other experiments, the preference data for DPO is constructed using a 16k subset from the dataset \textbf{ultrachat\_200k}. We have used $\theta_{SFT}$ to sample 32 or 64 responses per prompt, with $temperature = 1.0$ and $top\ p = 1.0$, which is consistent with the setup used in Nectar. During DPO and \model, we use the hyper-parameters the same as in Nectar.

\subsection{Experiment on HH-RLHF}

We perform SFT on $\text{HH-RLHF}_{SFT}$, following the previous hyper-parameters in Table~\ref{Nectar SFT hyper}.

When synthesizing the \model data, we filter the negative responses from 32 model responses with $a\%$ in $[100\%, 75\%, 50\%]$. The remaining hyperparameters are identical to those in Nectar.

When calculating the rewards for the test using \textbf{Starling-RM-7B-alpha}, for numerical stability, we added a layer of sigmoid function after the output of the model to map the reward values to $[0,1]$.

\end{document}